\documentclass{article}
\usepackage{spconf,amsmath,graphicx}
\usepackage{graphics}
\usepackage{xcolor}
\usepackage[normalem]{ulem}
\usepackage[top=2cm,bottom=2cm,left=2cm,right=2cm,a4paper]{geometry}
\usepackage{pstricks-add}
\usepackage{tikz}
\usepackage{subcaption}
\usepackage{booktabs}
\usepackage{multicol}
\usepackage{hyperref}
\DeclareRobustCommand\full  {\tikz[baseline=-0.6ex]\draw[thick] (0,0)--(0.5,0);}
\DeclareRobustCommand\dotted{\tikz[baseline=-0.6ex]\draw[thick,dotted] (0,0)--(0.5,0);}
\DeclareRobustCommand\dashed{\tikz[baseline=-0.6ex]\draw[thick,dashed] (0,0)--(0.5,0);}
\DeclareRobustCommand\chain {\tikz[baseline=-0.6ex]\draw[thick,dash dot] (0,0)--(0.5,0);}

\usepackage{xcolor}

\title{ON DESIGNING LIGHT-WEIGHT OBJECT TRACKERS THROUGH \\NETWORK PRUNING: USE CNNS or TRANSFORMERS?}
\name{\begin{tabular}{c}Saksham Aggarwal$^{\dagger *}$, Taneesh Gupta$^{\dagger *}$, Pawan K. Sahu$^{\dagger *}$, Arnav Chavan$^{\dagger}$, \\
Rishabh Tiwari$^{\dagger}$, Dilip K. Prasad$^{\ddagger}$ and Deepak K. Gupta$^{\dagger\ddagger}$\end{tabular}
\thanks{$^*$Equal Contribution}
\thanks{DKG received financial support from Texmin Foundation under grant number PSF-IH-IY-026. Work done while Saksham, Taneesh and Pawan were interns at Transmute AI Lab.}
}

\address{$^{\dagger}$Transmute AI Lab (Texmin Hub), Indian Institute of Technology (ISM) Dhanbad, India \\
$^{\ddagger}$Department of Computer Science, UiT The Arctic University of Norway, Tromso, Norway}

\begin{document}
\maketitle
%
%
%

%

%
\begin{abstract}
Object trackers deployed on low-power devices need to be light-weight, however, most of the current state-of-the-art (SOTA) methods rely on using compute-heavy backbones built using CNNs or Transformers. Large sizes of such models do not allow their deployment in low-power conditions and designing compressed variants of large tracking models is of great importance. This paper demonstrates how highly compressed light-weight object trackers can be designed using neural architectural pruning of large CNN and Transformer based trackers. Further, a comparative study on architectural choices best suited to design light-weight trackers is provided. A comparison between SOTA trackers using CNNs, Transformers as well as the combination of the two is presented to study their stability at various compression ratios. Finally results for extreme pruning scenarios going as low as 1\% in some cases are shown to study the limits of network pruning in object tracking. This work provides deeper insights into designing highly efficient trackers from existing SOTA methods.\footnote{Code is publicly available at \href{https://github.com/transmuteAI/Light-Weight-Trackers}{https://github.com/transmuteAI/Light-Weight-Trackers}}
\end{abstract}
\section{Introduction}
\label{sec:intro}




A general consensus about object tracking is that those using deeper convolutional neural network (CNN) architectures help in better discrimination of the target from the background and this is verified in siamese tracking networks. SiamFC \cite{bertinetto2016fully} uses an AlexNet-variant backbone whereas SiamRPN++ \cite{li2019siamrpn++} and SiamRCNN \cite{voigtlaender2020siam} use ResNet50 and ResNet101 \cite{he2016deep} backbones. As expected, the performance of SiamRPN++ is significantly better than SiamFC and SiamRCNN outperform the other two. Further in this landscape, we have deep Transformer models such as OSTrack \cite{ye2022joint} and STARK \cite{yan2021learning} that outperform all the CNN trackers stated above. While deeper models, be it CNN-based or Transformer-based \cite{vaswani2017attention, dosovitskiy2020image}, are found to help deliver better performance, they are generally large in size and often slow in terms of inference speed. This can be a big setback, especially for applications where close to real-time processing of the video frames is desired. There exist cases where beyond good accuracy, inference time on deployment device is of more importance.

In this paper, we focus on building light-weight object trackers that can be deployed on small devices and can operate in low-power conditions. Previous work on similar lines includes \cite{borsuk2021fear}, \cite{che2018channel}, and \cite{yan2021lighttrack}.
For fast and light-weight trackers, the existing dense trackers need to be compressed, and it is of interest to study how CNNs and transformers behave in such scenarios. To the best of our knowledge, our work is the first to identify which tracker model to choose for compression when designing a light-weight tracker. Beyond accuracy and inference speed, stability in the performance of the model is also important when building light trackers. It is of great interest to put all the SOTA trackers under the common roof of model size and the associated inference compute and study the drop in their performance for extreme model compression.

\textit{\textbf{Contributions}} (1) We present a comparative study on how the performance of SOTA object tracking methods is affected by different compression ratios. For this purpose, we choose three popular tracking methods: a fully CNN-based (Super-DiMP \cite{bhat2019learning}), a fully Transformer-based (OSTrack \cite{ye2022joint}), and a hybrid CNN-Transformer model (STARK \cite{yan2021learning}). (2) We study the effect of model compression on each tracker independently, as well as perform a comparative study across them for different budgets of inference computing. (3) For the robustness study, we analyze the inference of the compressed models on multiple tracking datasets and study the precision and success curves. (4) For the trackers to be really light-weight, the demanding level of compression is really high, and this implies that we need to push for extremely low inference budgets.  In this regard, we also study the stability of the models at inference budgets of as low as 1\% of the original base model.

\section{Pruning in Tracking}

\subsection{Background}
Model pruning \cite{tiwari2021chipnet,chavan2022vision,liu2017learning,chen2021chasing,frankle2018lottery} refers to inducing sparsity into the model architecture through removing the undesired or less important weights or filters. We present here a $pruning$-$in$-$tracking$ approach that can shrink their heavy CNN or Transformer backbones to reduce the number of channels, thereby improving their inference speed. We compress the model using a three-stage pipeline: 1) training the object tracking model with sparsity parameters and an added sparsity penalty function, 2) pruning the less important channels based on the learnt sparsity pattern, \hbox{and 3) fine-tuning the model to recover its performance.}

It has been observed that the post-pruning results are better if the pruning-related hyperparameters are embedded and the model is re-trained with them. Due to this reason, we train the network from scratch using the sparsity parameters. In general, the addition of these scaling parameters is not expected to affect the model performance since during the first training phase, it only serves to scale the weights. However, the added parameters need to be pushed towards 0 or 1 as a part of the training process. This requires adding an additional penalty on the intermediate values. For example, network slimming \cite{liu2017learning} uses $L_1$ norm penalty on the weights, and ChipNet \cite{tiwari2021chipnet} achieves it through Heaviside projections and added constraints. 



\textit{Learning sparsity masks}:
The core of our pruning-in-tracking method lies in how well the sparsity of masks is learnt. For this, we employ network slimming\cite{liu2017learning} directly leveraging $\gamma$ parameters of the batch-normalization layers as scaling factors. Since Transformers do not contain batchnorm layers, we prune MHSA and MLP modules by explicitly introducing masks as in \cite{chavan2022vision}. The scaling factors and the weights of the original backbone network are jointly trained with an additional sparsity regularization introduced on the scaling factors. Based on this, the optimization problem solved during training can be stated as
\begin{equation}
    \underset{\mathbf{W}, \boldsymbol\gamma}{\text{min}} \enskip \mathcal{L}(f(\boldsymbol\gamma \odot \mathbf{H}(\mathbf{w}); \mathbf{x}), \mathbf{y}) + \lambda g(\boldsymbol\gamma)
\label{eq-prune}
\end{equation}
where $(\mathbf{x}, \mathbf{y})$ denotes the training dataset, $\mathbf{H}$ denotes set of hidden channels, $\mathbf{w}$ denotes the weights of the network, and $\gamma_i \in \boldsymbol\gamma$ denotes the scaling factor associated with feature map/channel $\mathbf{c}_i \in \mathbf{H}$. Further, $\mathcal{L}(\cdot)$ denotes the standard training loss of the object tracking problem and $g(\cdot)$ denotes the sparsity-related penalty added to the scaling terms, weighted by hyper-parameter $\lambda$. Similar to \cite{liu2017learning}, we use $L_1$-norm for achieving sparsity.

\textit{Channel pruning and fine-tuning:}
\label{sec-prune-ft}
After learning the sparsity, the trained values of $\boldsymbol\gamma$ are used to remove the undesired set of channels from the network. There are many values in the distribution of $\gamma$ which are close to zero. To prune, we remove all the incoming and outgoing connections and the corresponding weights for a certain $\gamma_i$, and this can be achieved in a budget-aware manner. In \cite{liu2017learning}, channels are pruned with a global threshold across all layers. However, we have observed that maintaining the budget at a global level is generally not very efficient in extreme pruning scenarios. It leads to excessive pruning in the early layers and introduces a bottleneck on the information that can be propagated in the later layers. To circumvent this issue, we impose channel-budget separately at each layer or at the global level depending upon the budget and empirical performance. In the case of the former, the same value of channel-budget was used for every layer in the network, ensuring that the budget is satisfied at the global level.

\begin{figure*}
\centering
\small
\begin{subfigure}{1\textwidth}
\centering
\includegraphics[scale=.055]{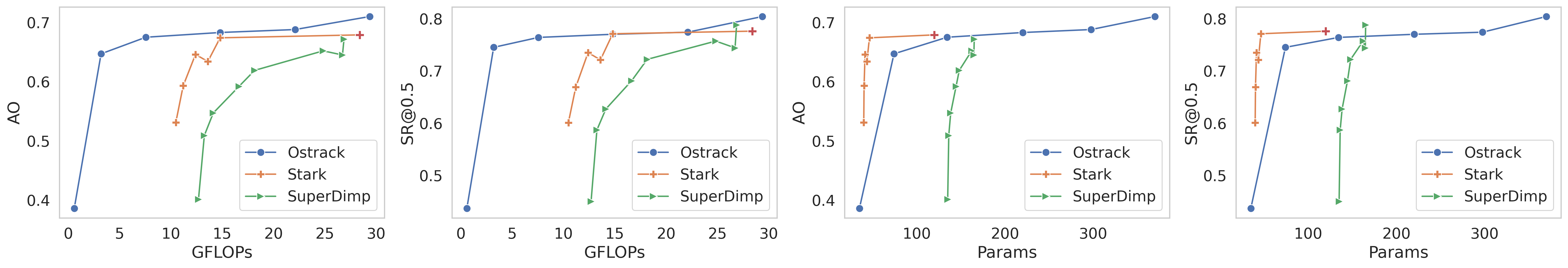}
\vspace{-1em}
\caption{GOT-10k}
\end{subfigure}
\begin{subfigure}{1\textwidth}
\centering
\includegraphics[scale=.055]{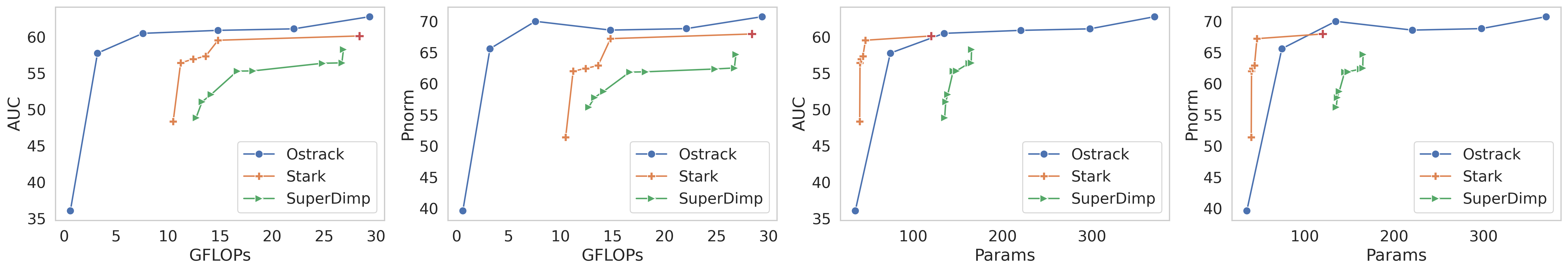}
\vspace{-1em}
\caption{LaSOT}
\end{subfigure}
\caption{Performance scores for for OSTrack \protect\cite{ye2022joint}, STARK \protect\cite{yan2021learning} and SuperDiMP \protect\cite{bhat2019learning} at different budgets of FLOPs and number of parameters in MB for two benchmark datasets.} 
\label{fig-results1}
\end{figure*}

\subsection{Light-weight SiamFC: A basic example}
We demonstrate the construction of light-weight object trackers through network pruning on SiamFC, a simple yet very effective object tracking method \cite{bertinetto2016fully}. This method forms the foundation for most current SOTA trackers, and a hypothesis experimented on this method should be transferable to the recent methods with minimal modifications. SiamFC uses a variant of AlexNet \cite{krizhevsky2017imagenet} as the backbone and the goal of the presented experiments is to perform network pruning over it and measure the performance vs. compute operations/model parameters tradeoff. 


\textbf{Implementation details. }SiamFC uses AlexNet-type backbone with five convolutional blocks.
Since the last layer does not use batch-normalization, we restrict the pruning of the model to only the first four convolutional layers. A sparsity penalty term, as indicated in Eq. \ref{eq-prune}, is used on all batch-normalization layers to learn the importance of the channels. The tracking model is then trained from scratch on the GOT-10k dataset \cite{huang2019got}. The hyperparameter settings for this stage are similar to those reported in the official implementation\footnote{Accessed at https://github.com/huanglianghua/siamfc-pytorch}.

Once the model is trained, channels corresponding to low values of the sparsity parameter are pruned. We report scores for a variety of channel budgets which denote the percentage of channels in the pruned network as compared to the parent network.  As motivated in Section \ref{sec-prune-ft}, we do not impose a global threshold across all layers of the network, rather a threshold is separately chosen for each layer in such a manner that the overall channel budget is satisfied. Unlike the results reported in \cite{liu2017learning}, we have observed that layerwise thresholding is more stable and better for pruning for this particular task.

\begin{table}
\small
    \centering
    \caption{Performance scores for various pruned variants of SiamFC tracking model at different channel budgets on GOT10k dataset. Here, `AO', `SR' and 'Params' stand for average overlap, success rate and parameters in MB, respectively. }
    \begin{tabular}{ r |  r | r | r | r | r}
         \textbf{Budget} & \textbf{AO} & \textbf{SR@50} & \textbf{SR@75} & \textbf{GFLOPs} & \textbf{Params} \\
         \midrule
          - & 0.343 & 0.377 & 0.107 & 2.812 & 8.93\\
          0.75 & 0.346 & 0.376 & 0.105 & 1.75 & 5.45  \\
          0.625 & 0.347 & 0.375 & 0.109 & 1.24 & 3.73 \\
         0.50 & 0.344 & 0.374 & 0.107 & 0.86 & 2.6 \\
         0.375 & 0.322 & 0.347 & 0.100 & 0.56 & 1.64 \\
         0.25 & 0.316 & 0.345 & 0.099 & 0.31 & 0.93 \\
         \bottomrule
    \end{tabular}
    \vspace{-5mm}
    \label{table-siamfc-results}
\end{table}

\textbf{Results. }Table \ref{table-siamfc-results} presents the results for the different pruning budgets. Across budgets, there is a small change in the observed tracking performance, however, it is almost negligible. For 75\% channel budget, there is no performance drop, however, the compute operations and model size are already reduced by around 40\% each. For 50\% budget, the gains are even higher (FLOPs reduction by 60\% and model size by 70\%) with no performance drop. A similar observation is made at higher compression levels, but with a minimal performance drop. For the extreme case of 25\% channel budget, the performance drops by only 8\% while compute FLOPs and model size reduce remarkably by 90\%. Depending upon the available hardware compute, the above experiments demonstrate the available control over the performance vs. speed trade-off for SiamFC.

\section{Light-weight deep trackers}
\vspace{-.8mm}
We have demonstrated that current object tracking methods can be compressed significantly to design light-weight trackers. Here, we investigate further the effect of compressing the SOTA deep trackers, and for this purpose, we choose CNN-based SuperDiMP \cite{bhat2019learning}, Transformer-based OSTrack \cite{ye2022joint} and a hybrid CNN-Transformer architecture STARK \cite{yan2021learning}. The choice of trackers covers a wide spectrum of well-performing architectural choices in computer vision, thus making this study thorough. 

SuperDiMP is a pure CNN-based tracker with bounding box regressor and classification head as well as a ResNet50 backbone. We prune the backbone as the structure of other components is not suitable for structured pruning. We implement pruning on the block level for ResNet-based SuperDiMP backbone as it gives superior results as compared to imposing a global budget. OSTrack uses attention layers throughout the network allowing to implicitly reduce FLOPs proportional to the pruning budget as shown in \cite{chavan2022vision}. Since our focus is on efficiency, we consider the smaller $256\times256$ image size in all our experiments for OSTrack. To prevent catastrophic pruning at extremely low inference budgets, we impose layerwise pruning at 1\% budget. Finally, experiments on encoder-decoder based STARK are presented. STARK requires a large number of epochs to train, and this makes it computationally infeasible for pruning experiments. Along with this, the FLOPs and Parameters in STARK are an order of magnitude higher than other trackers making a fairer comparison difficult. Hence, we designed a lighter version of this model, referred occasionally to as mini-STARK (4 stacks of encoder-decoder with 96 dimensions, 4 heads and a feedforward with 768 units), and this model provides performance similar to STARK, while the FLOPs are reduced by almost 2$\times$ and parameters by $3\times$. Even for mini-STARK, the use of global budget constraint leads to empty decoder layers and to circumvent this, we use layerwise budgets for this method. We also decoupled training and pruning of the encoder and decoder as coupling makes the decoder unstable. All the trackers are trained on GOT-10k dataset from scratch and tested on various datasets.


\begin{figure}
\small
\vspace{-2em}

\includegraphics[width=8cm, height=3.5cm]{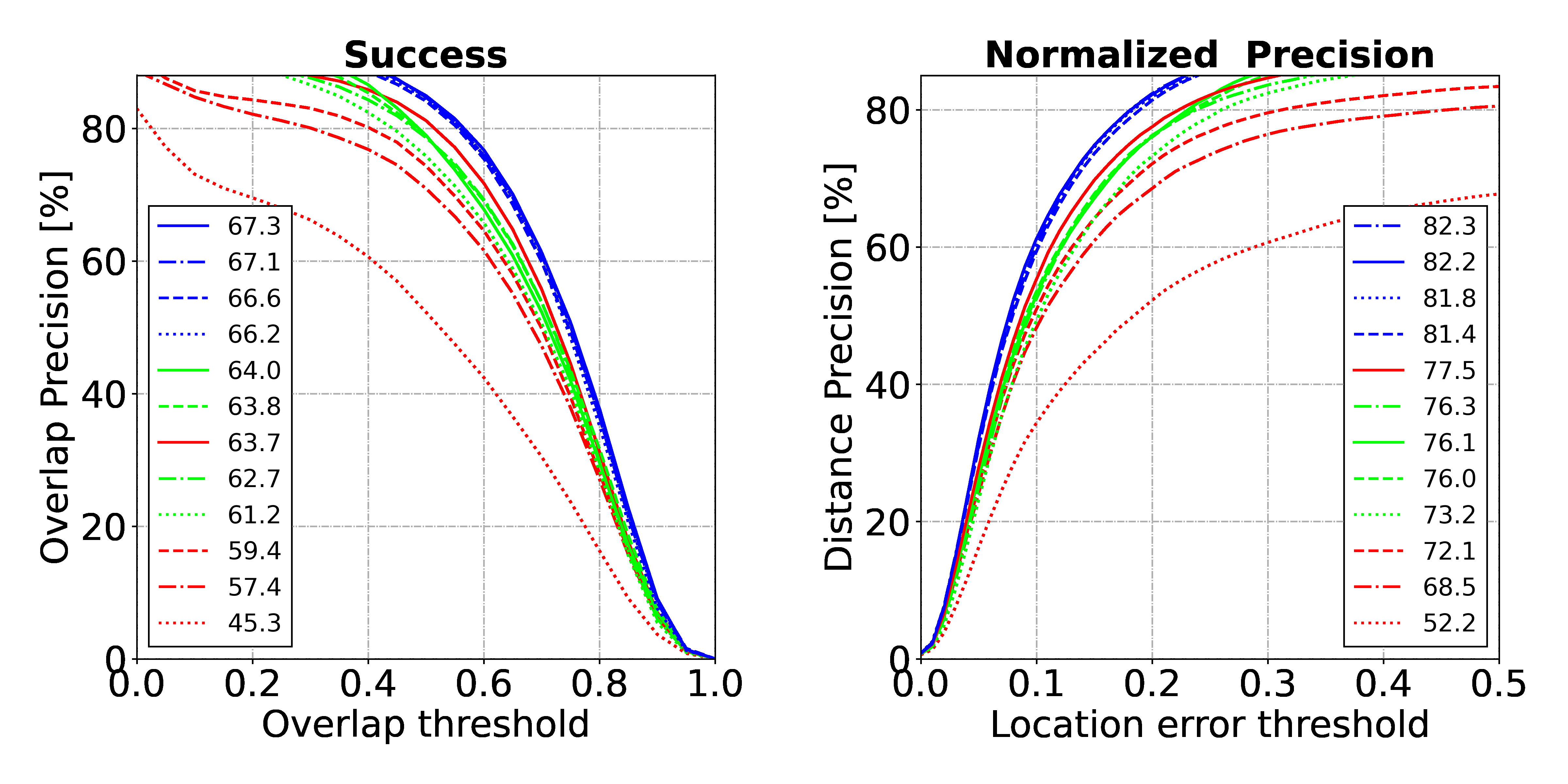}
\caption{Performance Plots. OSTrack ({\color{blue} \full}base, {\color{blue} \dashed}0.5, {\color{blue} \chain}0.25, {\color{blue} \dotted}0.1); mini-STARK ({\color{red} \full}base, {\color{red} \dashed}0.5, {\color{red} \chain}0.25, {\color{red} \dotted}0.1); SuperDiMP ({\color{green} \full}base, {\color{green} \dashed}0.5, {\color{green} \chain}0.25, {\color{green} \dotted}0.1).}
\vspace{-2mm}
\label{fig-otb}
\end{figure}

\begin{figure}
\centering
\small
\includegraphics[scale=.2205]{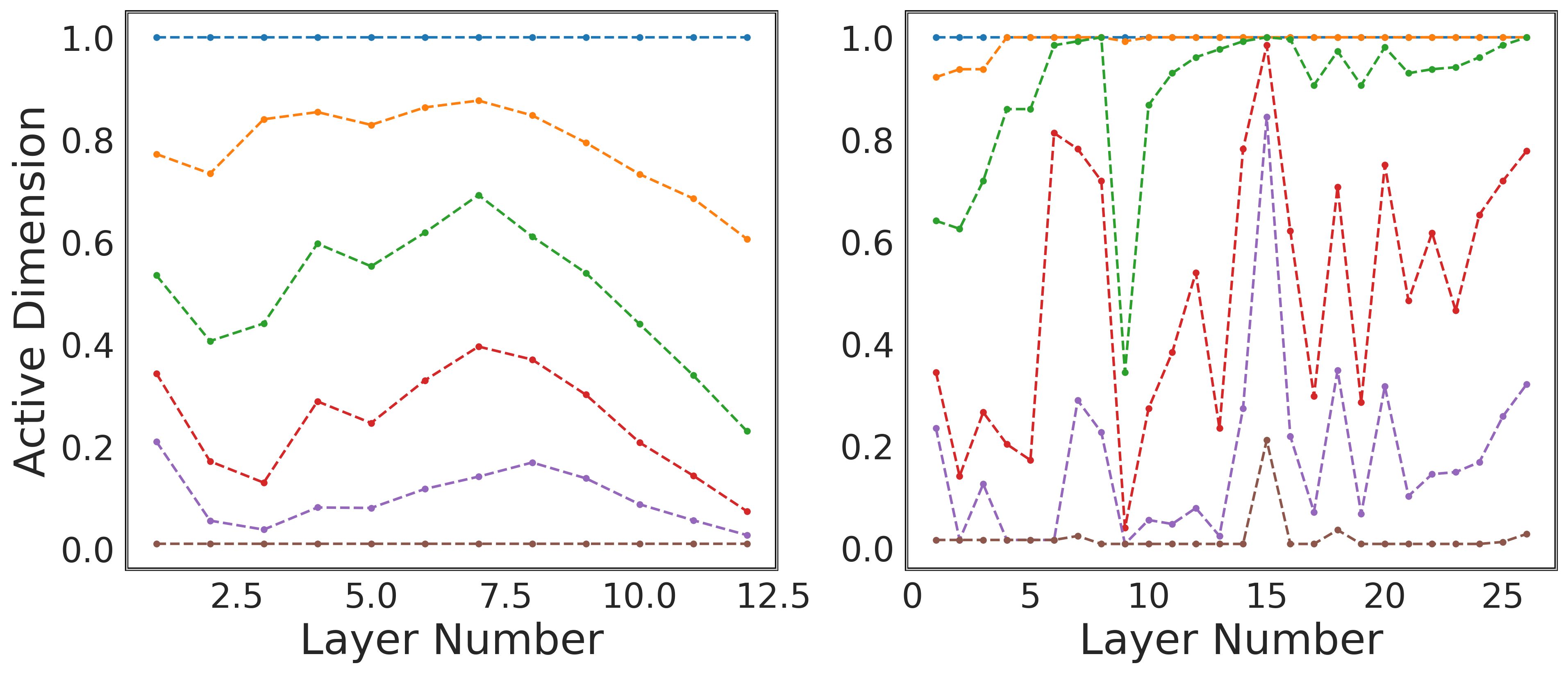}
\caption{\% Active dimensions at  different budgets for OSTrack (left) and SuperDiMP (right); Budgets: 1.0 ({\color{blue} \dashed}), 0.75 ({\color{orange} \dashed}), 0.5 ({\color{green} \dashed}), 0.25 ({\color{red} \dashed}), 0.1 ({\color{violet} \dashed}), 0.01 ({\color{brown} \dashed})}
\vspace{-6mm}
\label{fig-mlp-pattern}
\end{figure}

\subsection{Experimental insights}

Our experiments are based on channel pruning with channel budgets ranging from 75\% to even 1\% in some cases.

\begin{figure*}
\centering
\small
\begin{minipage}{0.67\textwidth}
\centering
\includegraphics[scale=.34]{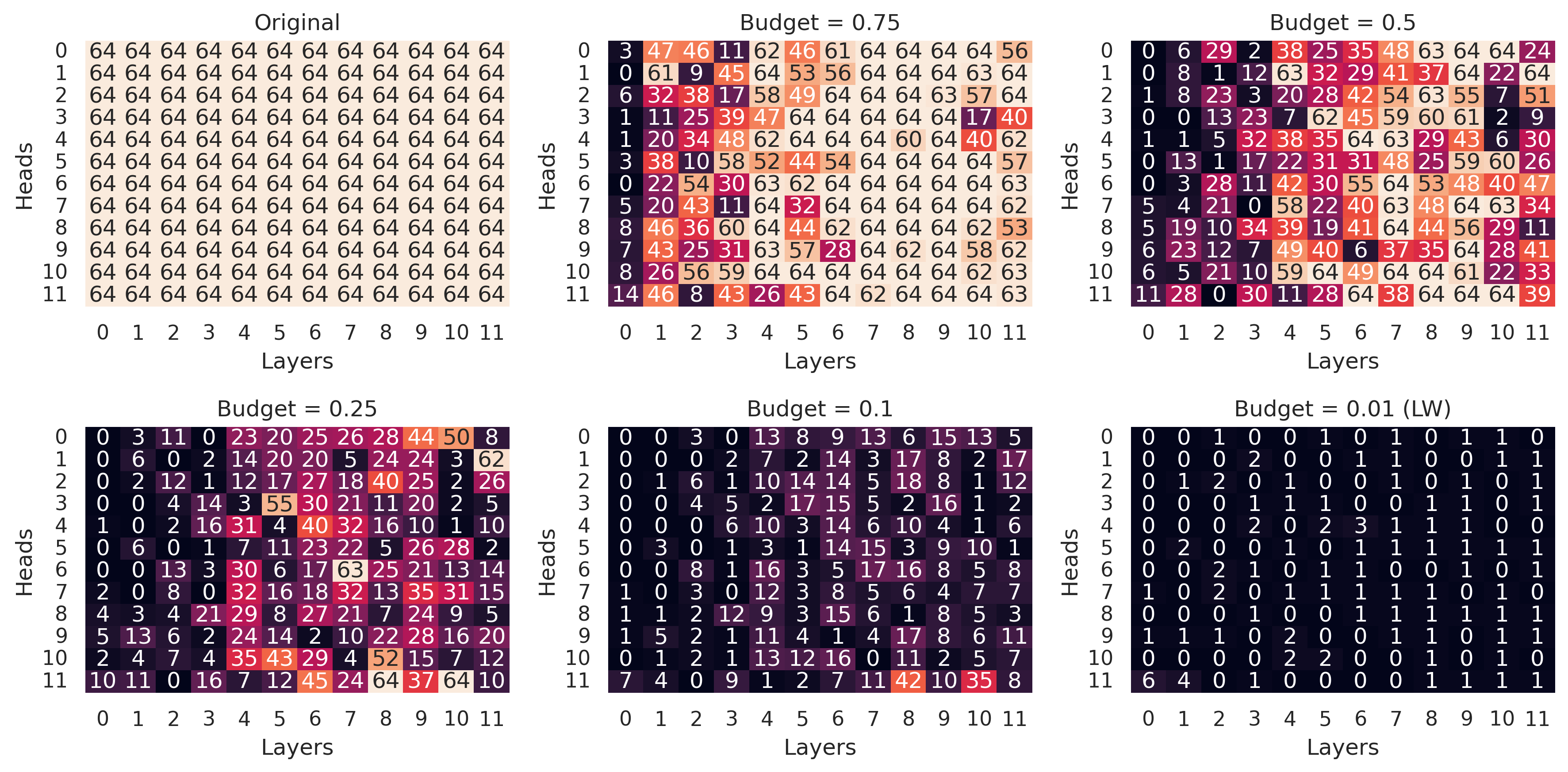}
\end{minipage}
\begin{minipage}{0.3\textwidth}\centering
\caption{Active attention dimensions per layer of OSTrack for different compression budgets.  All budgets are global, except the case of 0.01, which is layerwise. Initial attention modules are prioritized for removal for early compression budgets. For high compression levels, deeper and deeper attention modules are removed, however, in general, the network prefers to retain the later ones.}
\label{ost-attn}
\end{minipage}
\vspace{-1em}
\end{figure*}

\textbf{Performance of compressed models. }We first study the performance of the three trackers at different levels of compression. Fig. \ref{fig-results1} shows the performance comparisons for GOT-10k \cite{huang2019got} and LaSOT \cite{fan2019lasot} datasets. For all the trackers, the rightmost point on the curve corresponds to the full-scale models with no compression. For STARK, the performance of the original fullscale model is shown in red whereas mini-STARK corresponds to the second rightmost points in the plots of Fig. \ref{fig-results1}. Although the three trackers differ significantly in terms of the underlying architecture, the FLOPs are approximately the same with those for DiMP being slightly lower. However, OStrack is heavily overparameterized using more than 2$\times$ parameters than the other trackers, thereby adding to the memory overhead. 

A generic observation across the two datasets is that OSTrack outperforms the other trackers for almost all scenarios of compression. For an extremely less number of parameters (equivalent to approximately 1\% of the channels of the base model), OSTrack seems to be slightly unstable with STARK performing marginally better. Below 10 GFLOPs computation, the performance of STARK and SuperDiMP models significantly deteriorates. However, at this extreme compression as well, OSTrack is very stable with its performance being only marginally lower than the full-scale model. We see an interesting anomalous behavior of SuperDiMP on GOT-10k, where its full-scale configuration seems to perform better, however, it is only stable for this configuration and is outperformed at all compression budgets. Moreover, no such anomaly exists for LaSOT, which confirms SuperDiMP is not to be preferred.

We further perform a comparative study on OTB100 dataset \cite{wu2015object} (see Fig. \ref{fig-otb}). It is observed that OSTrack consistently outperforms the other two for all cases of channel budget. Further, the full-scale version of mini-STARK seems to perform better than SuperDiMP, however, this model is relatively less stable in terms of compression and the performance of the pruned versions drops significantly for the success as well as precision scores.

\textbf{Early compression dip. }A general observation across all cases of pruning is that during the initial pruning, the performance of all models dips, however, it is more prominent in SuperDiMP and mini-STARK. In Fig. \ref{fig-results1}, the two points around the dip correspond to the channel budgets of 75\% and 50\%. This dip occurs because the network finds it easier to remove the $1 \times 1$ channels from the network to achieve a certain budget. This is also confirmed by the fact that for 75\% budget, there occurs only a minimal drop in FLOPs. However, when pruned beyond, the network starts removing $3 \times 3$ filters as well and this arrangement is more optimal for the performance of the network, leading to improved generalization and an increase in its performance. This is also reflected by the bigger drop in FLOPs at this stage. Overall this initial dip is found to be common across all trackers that use CNNs, being more prominent for deep trackers. Although very small, a similar trend can also be found in the previous results shown for SiamFC.

\textbf{Induced sparsity pattern. }Next, we attempt to study the sparsity pattern induced in different trackers due to pruning. Fig. \ref{fig-mlp-pattern} shows the number of active dimensions per layer of OSTrack and SuperDiMP at different compression budgets. Note that for STARK, we perform layerwise pruning due to its instability issues, and the budgets are satisfied per layer. Clearly, a constant pattern would be observed in such a scenario and is not of interest to investigate further. For SuperDiMP, we do not perform any pruning at the input and output of each convolutional block. This is done to avoid inconsistency due to additional input head from the residual skip connections.

From Fig. \ref{fig-mlp-pattern}, we see that the active dimensions across different MLP layers of OSTrack are almost similar with higher pruning in the deeper layers of the network. For SuperDiMP, the trend is completely different. For its active dimensions, measured in terms of the fraction of active channels per layer, we see that more units are active in the middle of the network, however, the trend is not as smooth as OSTrack. For the attention modules of OSTrack as shown in Fig. \ref{ost-attn}, we see that for initial levels of compression, the starting layers of the network are preferred, however, the deeper and deeper layers get chosen for pruning as the compression level increases. This implies that the mixing of information through attention is not very relevant for the network during the initial layers and an adaptive choice is to retain them more towards the later parts, this is in contrast to classification models as shown in \cite{chavan2022vision}. For STARK, due to layerwise pruning, active units are the same across all layers, hence not reported here.

Overall, based on the experiments above, OSTrack is consistently better than the other trackers for almost all scenarios. It is stable at even extreme pruning levels and the performance drop is not very large. Further, since OSTrack is fully Transformer-based, imposing budget constraint on channels translates directly into FLOPs and memory budgets, and this provides a better control on the compression levels.
\vspace{-5mm}

\section{Conclusion}
\vspace{-3mm}
In this paper, we studied how light-weight object trackers can be designed by compressing the current SOTA deep trackers. First, we use a basic object tracking model which can be pruned significantly with a very limited drop in its performance. Further, we presented a comparative study on building light-weight trackers from SOTA tracking methods. We compared trackers that use CNNs, transformers as well as the combination of the two, and studied their stability at various compression levels, including extreme compression levels. Indicative in the presented experiments, a fully transformer-based model (OSTrack) exhibits both - better performance as well as better stability, when compared to the other two tracker models.

\newpage
\bibliographystyle{IEEEbib}
\bibliography{egbib}
\onecolumn

\appendix
\leftline{ {\Large Appendix } }
\section{Training Procedure}
\subsection{Sparsity Training}
All the trackers are initialized with weights pretrained on GOT10k dataset for early convergence. They are then further trained with induced sparsity.

\textbf{OSTrack} was trained with batch size of 128 at initial learning rate of 1$e$-4 for 256 image size and 1.25$e$-4 for 384 image size using AdamW optimizer and weight decay 1$e$-4. We use step learning rate strategy to decay learning rate by 0.1 after 40 epochs. The total training epochs are set to 100. Sparse Attention loss and sparse MLP loss are given a weightage of 1.5$e$-4 and 7.5$e$-5, respectively.

\textbf{SuperDiMP} was trained using Adam optimizer with learning rate decay of 0.2 every 15th epoch. The training epochs are set to 50, sampling 20,000 videos per epoch. Initial learning rates for classifier, bounding box regressor and feature extractor were kept 5$e$-5, 1$e$-3 and 2$e$-5 respectively. Scaling factor of 1$e$-4 was used for sparsity training. 

\textbf{Mini-STARK} is the lighter version of STARK (details are given in Section-3 of main paper). It is first trained from scratch for 300 epochs (since this is a new version of STARK from original). It was then trained with a learning rate of 5$e$-5 using AdamW optimizer and weight decay 1$e$-4. It was trained for 50 epochs. Sparse Attention loss and sparse MLP loss are given a weightage of 8$e$-4 and 1.5$e$-4, respectively for encoder while 7$e$-7 and 4.5$e$-7, respectively in case of decoder.

\subsection{Pruning}
In this stage, we create child networks according to a fixed budget.

In {\bf Ostrack}, we make a child network by using a global budget across MLP and Attention layers. In case of extreme pruning (1\% budget), layerwise pruning gave better results.

For {\bf SuperDiMP}, we present results (section-\ref{add_res}) on two pruning strategies: Blockwise and BNwise. Unlike in siamfc which constrains num-channels of the whole backbone to maintain budget,  we constrain channel pruning to maintain num-channel budget at each block level (Blockwise) or at each batchnorm layer (BNwise). In case of extreme pruning, it may result in a dead network due to complete removal of all channels from a layer. Therefore, minimum barrier was introduced to enforce a minimum number of 1\% channels or 1 channel, whichever is maximum, to remain in every layer after pruning.

For {\bf mini-STARK}, we use only layerwise pruning across all the budgets due to the issue of dead channels as explained in Section-3 of main paper.

\subsection{Fine Tuning}

The child networks obtained after pruning are then finetuned (initialized with sparsity train weights) using the same train settings as sparsity train but without the sparse parameters.

\subsection{Hardware}
All models are trained with NVIDIA A100 GPU (40GB).

\section{Additional Results} \label{add_res}

We experimented with different pruning strategies, namely Global, Layerwise, Blockwise and BNwise, which are defined below:

\begin{itemize}
    \item Global Pruning: Channels are pruned according to a global threshold across all the layers.
    \item Layerwise Pruning: A common budget is set for each layer of the model and pruning is done at the layer level according to the corresponding threshold.
    \item Blockwise Pruning: Model is pruned according to the budget defined at block level.
    \item BNwise Pruning: This is the finest level of pruning strategy in which pruning is done by assigning a common budget at each batchnorm of the model architecture.
\end{itemize}
Former two strategies strategies are employed by OSTrack while STARK uses Layerwise pruning. SuperDiMP utilizes the remaining two strategies along with a 'minBarrier' which keeps a certain minimum number of channels active to avoid total collapse.

Below are the results of all the three trackers - OSTrack, SuperDiMP and STARK on 4 well-known datasets - GOT10k, TrackingNet, LaSOT and OTB. Note that the trackers were trained on GOT10k dataset only. Further, to illustrate transfer-learning capabilities of the trackers, evaluation results on LaSOT, OTB and TrackingNet in addition to GOT10k are presented. 

\textbf{OSTrack} is inferred on input image sizes of 256$\times$256 and 384$\times$384. The former requires approx 2.5$\times$ less FLOPs as compared to the latter. Trend of accuracy and stability across different budgets is similar for both the image sizes. As we decrease the budget, the FLOPs are reduced by the same proportion except for 1\% budget. For extreme pruning budgets, FLOPs get dominated by the unprunable tracker head. For 1\% budget, we use layerwise pruning. Results for the four datasets are shown in tables [\ref{ostrack_got},\ref{ostrack_tnet},\ref{ostrack_lasot},\ref{ostrack_otb}].

\begin{table}[h]
    \centering
    \caption{Performance scores for various pruned variants of OSTrack tracking model at different channel budgets on GOT10k dataset. Here, `AO' and 'SR' stand for average overlap and success rate, respectively. 'Method' denotes either the pruning strategy used or the stage after which the tracker was evaluated}
    \begin{tabular}{l|c|c|c|c|c|l}
         \textbf{Model} & \textbf{Budget} & \textbf{G-FLOPs} & \textbf{AO} & \textbf{SR@50} & \textbf{SR@75} & \textbf{Method}  \\
         \midrule
         Ostrack 256(base) & - & 29.37695 & 0.710 & 0.804 & 0.682  & Reproduced  \\
         \midrule
         Ostrack & - & 29.37695 & 0.684 & 0.775 & 0.647 & After Sparse Training \\
         \midrule
         Ostrack  & 0.75 & 22.1051 & 0.688 & 0.774 & 0.651 & Global \\
         \midrule
         Ostrack  & 0.50 & 14.83337 & 0.683 & 0.770 & 0.643 & Global \\
         \midrule
         Ostrack  & 0.25 & 7.561574 & 0.675 & 0.764 & 0.628 & Global \\
         \midrule
         Ostrack  & 0.10 & 3.199269 & 0.647 & 0.745 & 0.570 & Global \\
         \midrule
       
         Ostrack  & 0.01 & 0.5868710 & 0.386 & 0.437 & 0.158 & Layerwise \\
         \midrule
         \midrule
         
        
         Ostrack 384 (base) & - & 71.65444 & 0.737 & 0.832 & 0.708  & Reproduced \\
         \midrule
         Ostrack & - & 71.65444 & 0.707 & 0.793 & 0.678 & After Sparse Training
           \\
         \midrule
         Ostrack  & 0.75 & 53.96579 & 0.706 & 0.794 & 0.677 & Global \\
         \midrule
         Ostrack  & 0.50 & 36.27715 & 0.707 & 0.797 & 0.673 & Global
          \\
         \midrule
         Ostrack  & 0.25 & 18.58852 & 0.682 & 0.775 & 0.637 & Global
          \\
         \midrule
         Ostrack  & 0.10 & 7.97729 &  0.654 & 0.749 & 0.579 & Global
          \\
         \midrule
         Ostrack  & 0.01 & 1.62362 & 0.413 & 0.471 & 0.173 & Layerwise
        \\
         \bottomrule

    \end{tabular}
    \label{ostrack_got}
\end{table}

\begin{table}[h]
    \centering
    \caption{Performance scores for various pruned variants of OSTrack tracking model at different channel budgets on TrackingNet dataset. Here, `Pnorm' and 'P' stand for normalised precision and precision, respectively. 'Method' denotes either the pruning strategy used or the stage after which the tracker was evaluated}
    \begin{tabular}{l|c|c|c|c|c|l}
         \textbf{Model} & \textbf{Budget} & \textbf{G-FLOPs} & \textbf{Success} & \textbf{Pnorm} & \textbf{P} & \textbf{Method}\\
         \midrule
         Ostrack 256(base) & - & 29.37695 & 81.7 &
         86.38 & 79.40 & Reproduced \\
         \midrule
         Ostrack  & - & 29.37695 & 80.64 & 85.40 & 78.11 & After Sparse Training\\
         \midrule
         Ostrack  & 0.75 & 22.1051 & 80.88 & 85.47 & 78.33 & Global\\
         \midrule
         Ostrack  & 0.50 & 14.83337 & 80.98 & 85.66 & 78.66 & Global\\
         \midrule
         Ostrack  & 0.25 & 7.561574 & 80.18 & 84.73 & 76.52 & Global\\
         \midrule
         Ostrack  & 0.10 & 3.199269 & 77.57 & 82.41 & 73.20 & Global\\
         \midrule
       
         Ostrack  & 0.01 & 0.5868710 & 54.49 & 58.55 & 43.29 & Layerwise\\
         \midrule
         \midrule
         
        
         Ostrack 384 (base) & - & 71.65444 & 82.88 & 87.40 & 81.56 & Reproduced\\
         \midrule
         Ostrack & - & 71.65444 & 81.91 & 86.55 & 80.27 & After Sparse Training \\
         \midrule
         Ostrack  & 0.75 & 53.96579 & 81.68 & 86.35 & 80.00 & Global\\
         \midrule
         Ostrack  & 0.50 & 36.27715 & 81.30 & 86.07 & 79.70 & Global\\
         \midrule
         Ostrack  & 0.25 & 18.58852 & 80.13 & 85.03 & 77.68 & Global\\
         \midrule
         Ostrack  & 0.10 & 7.97729 & 77.75 & 83.15 & 74.39 & Global\\
         \midrule
         Ostrack  & 0.01 & 1.62362 & 55.16 & 58.92 & 44.11 & Layerwise\\
         \bottomrule

    \end{tabular}
    \label{ostrack_tnet}
\end{table}

\begin{table}[h]
    \centering
    \caption{Performance scores for various pruned variants of OSTrack tracking model at different channel budgets on LaSOT dataset. Here, 'AUC', `Pnorm', 'P' and 'OP' stand for accuracy, normalised precision, precision and overlap precision, respectively. 'Method' denotes either the pruning strategy used or the stage after which the tracker was evaluated}
    \begin{tabular}{l|c|l|c|c|c|c|c|l}
         \textbf{Model} & \textbf{Budget} & \textbf{G-FLOPs} & \textbf{AUC} & \textbf{Pnorm} & \textbf{P} & \textbf{OP50} & \textbf{OP75} & \textbf{Method}\\
         \midrule
         Ostrack 256(base) & - & 29.37695 & 62.75 & 70.72 & 66.42 & 74.10 & 56.77 & Reproduced\\
         \midrule
         Ostrack & - & 29.37695 & 
         61.11 & 68.80 & 64.47 & 72.09 & 55.43 & After Sparse Training\\
         \midrule
         Ostrack  & 0.75 & 22.1051 & 
         61.08 & 68.81 & 64.58 & 72.02 & 55.22 & Global\\
         \midrule
         Ostrack  & 0.50 & 14.83337 & 
         60.87 & 68.57 & 64.47 & 71.70 & 54.94 & Global\\
         \midrule
         Ostrack  & 0.25 & 7.561574 & 
         60.46 & 69.98 & 64.92 & 71.91 & 54.57 & Global\\
         \midrule
         Ostrack  & 0.10 & 3.199269 & 
         57.74 & 65.59 & 59.22 & 68.24 & 49.97 & Global\\
         \midrule
       
         Ostrack  & 0.01 & 0.5868710 & 
         36.06 & 39.61 & 28.91 & 39.77 & 16.39 & Layerwise\\
         \midrule
         \midrule
         
        
         Ostrack 384 (base) & - & 71.65444 &
         63.48 & 71.65 & 67.86 & 74.93 & 58.25 & Reproduced\\
         \midrule
         Ostrack & - & 71.65444 &
         62.91 & 70.65 & 66.83 & 73.94 & 57.18 & After Sparse Training\\
         \midrule
         Ostrack  & 0.75 & 53.96579 & 
         62.86 & 70.62 & 66.86 & 73.88 & 56.97 & Global\\
         \midrule
         Ostrack  & 0.50 & 36.27715 & 
         62.13 & 69.91 & 66.06 & 73.06 & 56.03 & Global\\
         \midrule
         Ostrack  & 0.25 & 18.58852 & 
         62.01 & 70.31 & 65.77 & 73.30 & 55.83 & Global\\
         \midrule
         Ostrack  & 0.10 & 7.97729 &  
         59.00 & 67.41 & 61.55 & 69.98 & 51.39 & Global\\
         \midrule
         Ostrack  & 0.01 & 1.62362  &
         38.53 & 43.79 & 32.90 & 44.21 & 18.61 & Layerwise\\
         \bottomrule

    \end{tabular}
    \label{ostrack_lasot}
\end{table}

\begin{table}[h]
    \centering
    \caption{Performance scores for various pruned variants of OSTrack tracking model at different channel budgets on OTB dataset. Here, 'AUC', `Pnorm', 'P' and 'OP' stand for accuracy, normalised precision, precision and overlap precision, respectively. 'Method' denotes either the pruning strategy used or the stage after which the tracker was evaluated}
    \begin{tabular}{l|c|l|c|c|c|c|c|l}
         \textbf{Model} & \textbf{Budget} & \textbf{G-FLOPs} & \textbf{AUC} & \textbf{OP50} & \textbf{OP75} & \textbf{{P}} & \textbf{Pnorm} & \textbf{Method}\\
         \midrule
         Ostrack 256(base) & - & 29.37695 &
         67.35 & 84.5 & 49.88 & 89.18 & 82.37 & Reproduced\\
         \midrule
         Ostrack & - & 29.37695 &  67.11 & 84.24 & 49.49 & 88.59 & 82.08 & After Sparse Training
         \\
         \midrule
         Ostrack  & 0.75 & 22.1051 & 66.83 & 83.06 & 49.72 & 88.51 & 81.79 & Global
         \\
         \midrule
         Ostrack  & 0.50 & 14.83337 & 66.88 & 83.33 & 47.96 & 87.60 & 80.76 & Global\\
         \midrule
         Ostrack  & 0.25 & 7.561574 & 65.90 & 83.61 & 48.95 & 88.30 & 81.93 & Global\\
         \midrule
         Ostrack  & 0.10 & 3.199269 & 38.21 & 83.55 & 47.00 & 87.44 & 81.59 & Global
         \\
         \midrule
       
         Ostrack  & 0.01 & 0.5868710 & 38.21 & 41.88 & 14.10 & 51.51 & 44.18 & Layerwise\\
         \midrule
         \midrule
         
        
         Ostrack 384 (base) & - & 71.65444 & 67.30 & 84.93 & 50.75 & 88.13 & 82.17 & Reproduced\\
         \midrule
         Ostrack & - & 71.65444  & 66.50 & 83.56 & 49.75 & 87.18 & 81.22 & After Sparse Training\\
         \midrule
         Ostrack  & 0.75 & 53.96579 & 66.91 & 83.90 & 50.27 & 87.97 & 81.60 & Global\\
         \midrule
         Ostrack  & 0.50 & 36.27715 &  66.64 & 84.12 & 49.35 & 87.36 & 81.36 & Global\\
         \midrule
         Ostrack  & 0.25 & 18.58852 & 67.06 & 84.71 & 50.21 & 88.58 & 82.34 & Global\\
         \midrule
         Ostrack  & 0.10 & 7.97729 &  66.23 & 84.49 & 48.50 & 87.62 & 81.85 & Global\\
         \midrule
         Ostrack  & 0.01 & 1.62362 & 44.90 & 52.87 & 17.69 & 63.11 & 54.08 & Layerwise\\
         \bottomrule

    \end{tabular}
    \label{ostrack_otb}
\end{table}
 
In \textbf{SuperDiMP}, other modules apart from CNN backbone including BBOX-regressor and CLS-head are not pruned as the model experiences a huge performance drop otherwise. Due to this issue, it becomes challenging to extremely prune the tracker. For instance, around 12.6 total G-FLOPs still remain even at 1\% budget. In the tables [\ref{dimp_got},\ref{dimp_lasot},\ref{dimp_otb},\ref{dimp_tnet}], two FLOPs are mentioned - Backbone G-FLOPs and Total G-FLOPs. The former comes from CNN backbone only, while the latter additionally includes FLOPs from other modules too (like CLS-head). Among the two layerwise pruning methods, Blockwise is used across all budgets, while in the cases of extreme pruning (budget$\leq$10\%), BNwise results are presented too. It can be observed that Blockwise performs better than BNwise across all budgets and datasets. On the other hand, BNwise exhibits slightly better FLOPs reduction than Blockwise.
\begin{table}[h]
    \centering
    \caption{Performance scores for various pruned variants of Super-DiMP tracking model at different channel budgets on GOT10k dataset. Here, `AO' and `SR' stand for average overlap and success rate, respectively. 'Method' denotes either the pruning strategy used or the stage after which the tracker was evaluated}
    \begin{tabular}{l|c|l|l|c|c|c|l}
         & & \textbf{Bkbone} & \textbf{Total} & 
         & & & \\
         \textbf{Model} & \textbf{Budget} & \textbf{G-Flops} & \textbf{G-Flops} & 
         \textbf{AO} & \textbf{SR@50} & \textbf{SR@75} & 
         \textbf{Method} \\
         \midrule
         SuperDimp50 (base)& - & 16.2924 & 26.8746 &
         0.672 & 0.788 & 0.593 &
         Reproduced \\
         \midrule
         SuperDimp50 & 0.75 & 16.1357 & 26.7178 &
         0.645 & 0.744 & 0.557 & 
         Blockwise\\
         \midrule
         SuperDimp50 & 0.50 & 14.2585 & 24.8407 &
         0.652 & 0.757 & 0.560 &
         Blockwise\\
         \midrule
         SuperDimp50 & 0.25 & 7.5743 & 18.1565 &
         0.619 & 0.722 & 0.499 &
         Blockwise\\
         \midrule
         SuperDimp50 & 0.20 & 6.0616 & 16.6438 &
         0.592 & 0.681 & 0.469 &
         Blockwise\\
         \midrule
         SuperDimp50 & 0.10 & 3.5442 & 14.1263 &
         0.547 & 0.627 & 0.411 &
         Blockwise + minBarrier\\
         \midrule
         SuperDimp50 & 0.05 & 2.6805 & 13.2627 &
         0.509 & 0.587 & 0.365 &
         Blockwise + minBarrier\\
         \midrule
         SuperDimp50 & 0.01 & 2.1161 & 12.6983 &
         0.401 & 0.450 & 0.243 &
         Blockwise + minBarrier\\
         \midrule
         SuperDimp50 & 0.10 & 2.6568 & 13.2390 &
         0.550 & 0.634 & 0.399 &
         BNwise\\
         \midrule
         SuperDimp50 & 0.05 & 2.2668 & 12.8490 &
         0.483 & 0.556 & 0.328 &
         BNwise\\
         \midrule
         SuperDimp50 & 0.01 & 1.9905 & 12.5727 &
         0.385 & 0.434 & 0.235 &
         BNwise + minBarrier\\
         \bottomrule

    \end{tabular}
    \label{dimp_got}
\end{table}

\begin{table}[h]
    \centering
    \caption{Performance scores for various pruned variants of Super-DiMP tracking model at different channel budgets on LaSOT dataset. Here, `AUC', 'Pnorm' and 'P' stand for accuracy, normalised precision and precision, respectively. 'Method' denotes either the pruning strategy used or the stage after which the tracker was evaluated}
    \begin{tabular}{l|c|l|l|c|c|c|l}
         & & \textbf{Bkbone} & \textbf{Total} & 
         & & & \\
         \textbf{Model} & \textbf{Budget} & \textbf{G-Flops} & \textbf{G-Flops} & 
         \textbf{AUC} & \textbf{Pnorm} & \textbf{P} &
         \textbf{Method} \\
         \midrule
         SuperDimp50 (base)& - & 16.2924 & 26.8746 &
         58.23 & 64.66 & 51.87 &
         Reproduced \\
         \midrule
         SuperDimp50 & 0.75 & 16.1357 & 26.7178 &
         56.40 & 62.50 & 56.11 &
         Blockwise\\
         \midrule
         SuperDimp50 & 0.50 & 14.2585 & 24.8407 &
         56.34 & 62.33 & 55.30 &
         Blockwise\\
         \midrule
         SuperDimp50 & 0.25 & 7.5743 & 18.1565 &
         55.29 & 61.88 & 54.20 &
         Blockwise\\
         \midrule
         SuperDimp50 & 0.20 & 6.0616 & 16.6438 &
         55.26 & 61.83 & 54.18 &
         Blockwise\\
         \midrule
         SuperDimp50 & 0.10 & 3.5442 & 14.1263 &
         52.04 & 58.74 & 52.04 &
         Blockwise + minBarrier\\
         \midrule
         SuperDimp50 & 0.05 & 2.6805 & 13.2627 &
         51.04 & 57.75 & 48.31 &
         Blockwise + minBarrier\\
         \midrule
         SuperDimp50 & 0.01 & 2.1161 & 12.6983 &
         48.86 & 56.20 & 44.65 &
         Blockwise + minBarrier\\
         \midrule
         SuperDimp50 & 0.10 & 2.6568 & 13.2390 &
         50.33 & 55.14 & 47.83 &
         BNwise\\
         \midrule
         SuperDimp50 & 0.05 & 2.2668 & 12.8490 &
         47.26 & 53.26 & 43.22 &
         BNwise\\
         \midrule
         SuperDimp50 & 0.01 & 1.9905 & 12.5727 &
         44.41 & 52.01 & 40.52 &
         BNwise + minBarrier\\
         \bottomrule

    \end{tabular}
    \label{dimp_lasot}
\end{table}

\begin{table}[h]
    \centering
    \caption{Performance scores for various pruned variants of Super-DiMP tracking model at different channel budgets on OTB dataset. Here, `AUC', 'Pnorm' and 'P' stand for accuracy, normalised precision and precision, respectively. 'Method' denotes either the pruning strategy used or the stage after which the tracker was evaluated}
    \begin{tabular}{l|c|l|l|c|c|c|l}
         & & \textbf{Bkbone} & \textbf{Total} & 
         & & & \\
         \textbf{Model} & \textbf{Budget} & \textbf{G-Flops} & \textbf{G-Flops} & 
         \textbf{AUC} & \textbf{Pnorm} & \textbf{P} &
         \textbf{Method} \\
         \midrule
         SuperDimp50 (base)& - & 16.2924 & 26.8746 &
         65.14 & 76.93 & 84.51 & 
         Reproduced \\
         \midrule
         SuperDimp50 & 0.75 & 16.1357 & 26.7178 &
         64.02 & 76.11 & 83.63 &
         Blockwise\\
         \midrule
         SuperDimp50 & 0.50 & 14.2585 & 24.8407 &
         63.82 & 76.01 & 83.62 &
         Blockwise\\
         \midrule
         SuperDimp50 & 0.25 & 7.5743 & 18.1565 &
         62.70 & 76.25 & 82.74 &
         Blockwise\\
         \midrule
         SuperDimp50 & 0.20 & 6.0616 & 16.6438 &
         63.12 & 75.57 & 82.66 &
         Blockwise\\
         \midrule
         SuperDimp50 & 0.10 & 3.5442 & 14.1263 &
         61.24 & 73.22 & 81.04 &
         Blockwise + minBarrier\\
         \midrule
         SuperDimp50 & 0.05 & 2.6805 & 13.2627 &
         59.59 & 71.66 & 78.42 &
         Blockwise + minBarrier\\
         \midrule
         SuperDimp50 & 0.01 & 2.1161 & 12.6983 &
         52.26 & 62.78 & 70.64 &
         Blockwise + minBarrier\\
         \midrule
         SuperDimp50 & 0.10 & 2.6568 & 13.2390 &
         58.37 & 70.18 & 76.63 &
         BNwise\\
         \midrule
         SuperDimp50 & 0.05 & 2.2668 & 12.8490 &
         55.85 & 67.13 & 73.49 &
         BNwise\\
         \midrule
         SuperDimp50 & 0.01 & 1.9905 & 12.5727 &
         49.89 & 60.25 & 67.91 &
         BNwise + minBarrier\\
         \bottomrule

    \end{tabular}
    \caption{SuperDiMP on OTB}
    \label{dimp_otb}
\end{table}

\begin{table}[h]
    \centering
    \caption{Performance scores for various pruned variants of Super-DiMP tracking model at different channel budgets on TrackingNet dataset. Here, 'Pnorm' and 'P' stand for normalised precision and precision, respectively. 'Method' denotes either the pruning strategy used or the stage after which the tracker was evaluated}
    \begin{tabular}{l|c|l|l|c|c|c|l}
         & & \textbf{Bkbone} & \textbf{Total} & 
         & & & \\
         \textbf{Model} & \textbf{Budget} & \textbf{G-Flops} & \textbf{G-Flops} & 
         \textbf{Success} & \textbf{Pnorm} & \textbf{P} &
         \textbf{Method} \\
         \midrule
         SuperDimp50 (base)& - & 16.2924 & 26.8746 &
         76.7 & 81.33 & 71.14 & 
         Reproduced \\
         \midrule
         SuperDimp50 & 0.75 & 16.1357 & 26.7178 &
         75.56 & 79.66 & 69.19 &
         Blockwise\\
         \midrule
         SuperDimp50 & 0.50 & 14.2585 & 24.8407 &
         75.16 & 79.08 & 68.89 &
         Blockwise\\
         \midrule
         SuperDimp50 & 0.25 & 7.5743 & 18.1565 &
         74.84 & 79.04 & 68.06 &
         Blockwise\\
         \midrule
         SuperDimp50 & 0.20 & 6.0616 & 16.6438 &
         73.62 & 78.09 & 66.52 &
         Blockwise\\
         \midrule
         SuperDimp50 & 0.10 & 3.5442 & 14.1263 &
         71.67 & 76.55 & 64.00 &
         Blockwise + minBarrier\\
         \midrule
         SuperDimp50 & 0.05 & 2.6805 & 13.2627 &
         69.26 & 74.01 & 60.95 &
         Blockwise + minBarrier\\
         \midrule
         SuperDimp50 & 0.01 & 2.1161 & 12.6983 &
         62.02 & 66.69 & 52.62 &
         Blockwise + minBarrier\\
         \midrule
         SuperDimp50 & 0.10 & 2.6568 & 13.2390 &
         71.28 & 75.74 & 63.33 &
         BNwise\\
         \midrule
         SuperDimp50 & 0.05 & 2.2668 & 12.8490 &
         69.34 & 73.68 & 60.21 &
         BNwise\\
         \midrule
         SuperDimp50 & 0.01 & 1.9905 & 12.5727 &
         61.85 & 66.21 & 51.70 &
         BNwise + minBarrier\\
         \bottomrule
         
    \end{tabular}
    \caption{SuperDiMP on TrackingNet}
    \label{dimp_tnet}
\end{table}

In \textbf{Mini-Stark}, only encoder-decoder transformer is pruned whereas other modules including CNN backbone and head are not pruned as the model experiences a huge performance drop otherwise. Due to this issue, it becomes hard to reduce FLOPs significantly. For instance, around 10.48 total G-FLOPs still remain even at 10\% budget which is much larger as compared to OSTrack and DiMP. In the tables [\ref{stark_got},\ref{stark_lasot},\ref{stark_otb},\ref{stark_tnet} results are provided for the above mentioned four datasets.
In MiniStark, layerwise pruning is used as naive pruning leads to more vanishing of neurons from the decoder layers which creates bottleneck and leads to marginal drop in evaluation performance across all object tracking datasets.

\begin{table}[h]
    \centering
    \caption{Performance scores for various pruned variants of STARK tracking model at different channel budgets on GOT10k dataset. Here, `AO' and `SR' stand for average overlap and success rate, respectively. 'Method' denotes either the pruning strategy used or the stage after which the tracker was evaluated}
    \begin{tabular}{l|c|l|c|c|c|l}
         \textbf{Model} & \textbf{Budget} & \textbf{G-FLOPs} & \textbf{AO} & \textbf{SR@50} & \textbf{SR@75} & \textbf{Method}\\
         \midrule
         Stark-ST50 (base) & - & 29.4105 & 0.679 & 0.776 & 0.622 & Reproduced \\
         \midrule
         mini-STARK ST50 & - & 14.8345 & 0.674 & 0.771 & 0.600 & Standard Implemetation\\
         \midrule
         mini-STARK ST50 & - & 14.8345 & 0.660 & 0.763 & 0.576 & After Sparse Training\\
         \midrule
         mini-STARK ST50 & 0.75 & 13.6251 & 0.634 & 0.721 & 0.552 & Layerwise\\
         \midrule
         mini-STARK ST50 & 0.5 & 12.4167 & 0.646 & 0.735 & 0.572 & Layerwise\\
         \midrule
         mini-STARK ST50 & 0.25 & 11.2578 & 0.593 & 0.669 & 0.504 & Layerwise\\
         \midrule
       
         mini-STARK ST50 & 0.10 & 10.4876 & 0.531 & 0.601 & 0.432 & Layerwise\\
         \bottomrule

    \end{tabular}
    \newline
    \label{stark_got}
\end{table}


\begin{table}[h]
    \centering
    \caption{Performance scores for various pruned variants of STARK tracking model at different channel budgets on TrackingNet dataset. Here, 'Pnorm' and 'P' stand for normalised precision and precision, respectively. 'Method' denotes either the pruning strategy used or the stage after which the tracker was evaluated}
    \begin{tabular}{l|c|l|c|c|c|l}
         \textbf{Model} & \textbf{Budget} & \textbf{G-FLOPs} & \textbf{Success} & \textbf{Pnorm} & \textbf{P} & \textbf{Method}\\
         \midrule
         Stark-ST50 (base) & - & 29.4105 & 79.47 & 83.87 & 76.05 & Reproduced \\
         \midrule
         mini-STARK ST50 & - & 14.8345 & 78.00 & 82.79 & 73.22 & Standard Implemetation\\
         \midrule
         mini-STARK ST50 & - & 14.8345 & 78.25 & 82.79 & 74.27 & After Sparse Training\\
         \midrule
         mini-STARK ST50 & 0.75 & 13.6251 & 75.79 & 80.35 & 71.41 & Layerwise\\
         \midrule
         mini-STARK ST50 & 0.5 & 12.4167 & 76.17 & 80.39 & 71.46 & Layerwise\\
         \midrule
         mini-STARK ST50 & 0.25 & 11.2578 & 74.92 & 78.94 & 69.7 & Layerwise\\
         \midrule
       
         mini-STARK ST50 & 0.10 & 10.4876 & 63.18 & 65.23 & 55.43 & Layerwise\\
         \bottomrule

    \end{tabular}
    \label{stark_tnet}
\end{table}

\begin{table}[h]
    \centering
    \caption{Performance scores for various pruned variants of STARK tracking model at different channel budgets on LaSOT dataset. Here, `AUC', `Pnorm' and 'P' stand for accuracy, normalised precision and precision, respectively. 'Method' denotes either the pruning strategy used or the stage after which the tracker was evaluated}
    \begin{tabular}{l|c|l|c|c|c|l}
         \textbf{Model} & \textbf{Budget} & \textbf{G-FLOPs} & \textbf{AUC} & \textbf{Pnorm} & \textbf{P} & \textbf{Method}\\
         \midrule
         Stark-ST50 (base) & - & 29.4105 & 60.07 & 68.03 & 61.86 & Reproduced \\
         \midrule
         mini-STARK ST50 & - & 14.8345 & 59.51 & 67.2 & 60.31 & Standard Implemetation\\
         \midrule
         mini-STARK ST50 & - & 14.8345 & 59.53 & 67.85 & 60.95 & After Sparse Training\\
         \midrule
         mini-STARK ST50 & 0.75 & 13.6251 & 57.31 & 62.88 & 55.89 & Layerwise\\
         \midrule
         mini-STARK ST50 & 0.50 & 12.4167 & 56.91 & 62.42 & 55.52 & Layerwise\\
         \midrule
         mini-STARK ST50 & 0.25 & 11.2578 & 56.40 & 61.98 & 55.09 & Layerwise\\
         \midrule
       
         mini-STARK ST50 & 0.10 & 10.4876 & 48.32 & 51.37 & 43.65 & Layerwise\\
         \bottomrule

    \end{tabular}
    \label{stark_lasot}
\end{table}

\begin{table}[h]
    \centering
    \caption{Performance scores for various pruned variants of STARK tracking model at different channel budgets on OTB dataset. Here, `AUC', `Pnorm', 'P' and 'OP' stand for accuracy, normalised precision, precision and overlap precision, respectively. 'Method' denotes either the pruning strategy used or the stage after which the tracker was evaluated}
    \begin{tabular}{l|c|l|c|c|c|c|c|l}
         \textbf{Model} & \textbf{Budget} & \textbf{G-FLOPs} & \textbf{AUC} & \textbf{Pnorm} & \textbf{P} & \textbf{OP50} & \textbf{OP75} & \textbf{Method}\\
         \midrule
         Stark-ST50 (base) & - & 29.4105 & 65.61 & 83.40 & 46.12 & 86.35 & 80.02 & Reproduced \\
         \midrule
         mini-STARK ST50 & - & 14.8345 & 63.74 & 81.19 & 44.72 & 83.96 & 77.49& Standard Implemetation\\
         \midrule
         mini-STARK ST50 & - & 14.8345 & 64.22 & 81.76 & 46.38 & 84.17 & 78.49 & After Sparse Training\\
         \midrule
         mini-STARK ST50 & 0.75 & 13.6251 & 58.54 & 72.83 & 39.65 & 76.42 & 70.74 & Layerwise\\
         \midrule
         mini-STARK ST50 & 0.5 & 12.4167 & 59.40 & 74.34 & 39.50 & 78.14 & 72.11 & Layerwise\\
         \midrule
         mini-STARK ST50 & 0.25 & 11.2578 & 57.44 & 70.94 & 37.89 & 74.85 & 68.52 & Layerwise\\
         \midrule
       
         mini-STARK ST50 & 0.10 & 10.4876 & 45.34 & 52.29 & 23.69 & 58.79 & 52.23 & Layerwise\\
         \bottomrule

    \end{tabular}
    \label{stark_otb}
\end{table}

\end{document}